\pdfoutput=1
\documentclass[11pt]{article}
\usepackage{acl}     

\usepackage{multirow}  % 
\usepackage{colortbl}  % 
\usepackage{booktabs}  % 
\usepackage{amsmath} % 
\usepackage{subcaption} % 
\usepackage{booktabs, multirow, siunitx} % 
\usepackage{stfloats}  % 

\usepackage[utf8]{inputenc}

\usepackage{algorithm}
\usepackage{algpseudocode}
\usepackage{float}

\restylefloat{algorithm}

\usepackage{natbib}      % 
\usepackage{graphicx}    %
\usepackage{enumitem}    % 
\usepackage{hyperref}    % 

\usepackage{times}
\usepackage{latexsym}

\usepackage[T1]{fontenc}
\usepackage[utf8]{inputenc}

\usepackage{microtype}

\usepackage{inconsolata}

\usepackage{graphicx}

\usepackage{authblk}

\title{Identifying Pre-training Data in LLMs: A Neuron Activation-Based Detection Framework}

\author{%
  \textbf{Hongyi Tang\textsuperscript{*}, Zhihao Zhu\textsuperscript{*}, Yi Yang}\\
  The Hong Kong University of Science and Technology\\
  \texttt{\{hongyitang,zhihaozhu,imyiyang\}@ust.hk}%
}

\date{}  % 不显示日期

\begin{document}
\maketitle
\begin{abstract}
The performance of large language models (LLMs) is closely tied to their training data, which can include copyrighted material or private information, raising legal and ethical concerns. Additionally, LLMs face criticism for dataset contamination and internalizing biases. To address these issues, the Pre-Training Data Detection (PDD) task was proposed to identify if specific data was included in an LLM's pre-training corpus. However, existing PDD methods often rely on superficial features like prediction confidence and loss, resulting in mediocre performance. To improve this, we introduce NA-PDD, a novel algorithm analyzing differential neuron activation patterns between training and non-training data in LLMs. This is based on the observation that these data types activate different neurons during LLM inference. We also introduce CCNewsPDD, a temporally unbiased benchmark employing rigorous data transformations to ensure consistent time distributions between training and non-training data. Our experiments demonstrate that NA-PDD significantly outperforms existing methods across three benchmarks and multiple LLMs.
\end{abstract}

\section{Introduction}
The effectiveness of large language models (LLMs) hinges significantly on their training corpus \cite{kaplan2020scaling, gao2020pile}. However, these pre-training corpora may contain copyrighted material \cite{chang2023speak, mozes2023use} or private user information \cite{yao2024survey, liu2021machine}, raising substantial concerns about compliance and privacy. For example, The New York Times recently filed a lawsuit against OpenAI, alleging illegal use of its articles as training data for ChatGPT \footnote{https://www.nytimes.com/2023/12/27/business/media/new-york-times-open-ai-microsoft-lawsuit.html}. Furthermore, LLMs can inadvertently acquire undesirable knowledge from their training data, such as biased \cite{ferrara2023should, kotek2023gender} or harmful content \cite{deshpande2023toxicity, gehman2020realtoxicityprompts}, compromising the trustworthiness of the language model. Precise knowledge of the learned data is therefore crucial. However, determining whether a model has incorporated specific data remains challenging. This leads to a critical question: \textit{given an LLM and a text sample, how can we determine if this text was part of the LLM's pre-training?} This is the pre-training data detection (PDD) problem.

Existing PDD algorithms suffer from two primary limitations: 1) \textbf{Superficial Information Reliance}: Most algorithms focus on surface-level features of LLMs \cite{carlini2023extracting, zhang2024min}. For instance, Loss Attack \cite{yeom2018privacy} uses the LLM's prediction loss on a given text, while Min-K\% Prob \cite{shi2023detecting} uses predictive probabilities of tokens. This approach limits detection effectiveness, resulting in insufficient performance and high false positive rates, rendering them unsuitable for applications such as copyright verification \cite{duan2024membership, zhang2024membership}. 2) \textbf{Benchmark Time Drift}: Due to the confidentiality of LLM training data \cite{achiam2023gpt, bai2023qwen}, researchers often use release dates to infer training data, comparing it to publicly available datasets like Wikipedia. For example, pre-2023 data might be considered training data for a 2023 LLaMA model, while post-2023 data is viewed as non-training data \cite{shi2023detecting}. This temporal bias complicates the accurate evaluation of PDD methods intended to identify training corpora.

To address the first limitation, we introduce \textbf{NA-PDD}, a novel PDD algorithm that utilizes neuronal activation patterns within LLMs. Our method stems from the observation that training text activates different neurons within an LLM compared to non-training text. NA-PDD is particularly suitable for copyright verification in open-source LLMs or internal audits. It utilizes a small set of reference corpora to record neuronal activation for both training and non-training data. Neurons predominantly activated by training data are labeled as "member" neurons, while those activated by non-training data are labeled as "non-member" neurons. We then design a straightforward detection algorithm to determine whether a given sample $x$ was part of the LLM's pre-training corpus. During model inference with input $x$, we record the activation states of neurons across different layers and provide PDD predictions based on the relative prominence of member neurons in these layers.

To address the second limitation, we introduce \textbf{CCNewsPDD}, a time-drift-free PDD benchmark based on the CCNews dataset. This benchmark ensures temporal alignment between training and non-training data. To make sure that non-training data was not used in pre-training, we apply transformations such as back translation, masking, and LLM rewriting to the original non-training data. These transformations introduce meaningful variations while maintaining a rigorous definition of non-training data.

Our contributions are as follows:
\begin{itemize}
    \item We introduce NA-PDD, a novel PDD algorithm leveraging neuronal activation patterns within LLMs. NA-PDD analyzes the differential activation between trained and non-trained samples during inference to construct an effective PDD algorithm.
    \item We introduce CCNewsPDD, a time-drift-free PDD benchmark. Using data transformation methods, CCNewsPDD ensures no temporal distribution differences between training and non-training data while maintaining semantic and lexical coherence.
    \item We evaluate NA-PDD against nine representative PDD methods on CCNewsPDD and two public benchmarks. Our results demonstrate substantial improvements. For example, on OPT-6.7B with CCNewsPDD, NA-PDD outperforms DC-PDD by 27.9\% AUC points (increasing from 71.8\% to 99.7\%).
\end{itemize}

\begin{figure*}[!ht]   
  \centering
  \includegraphics[width=\textwidth]{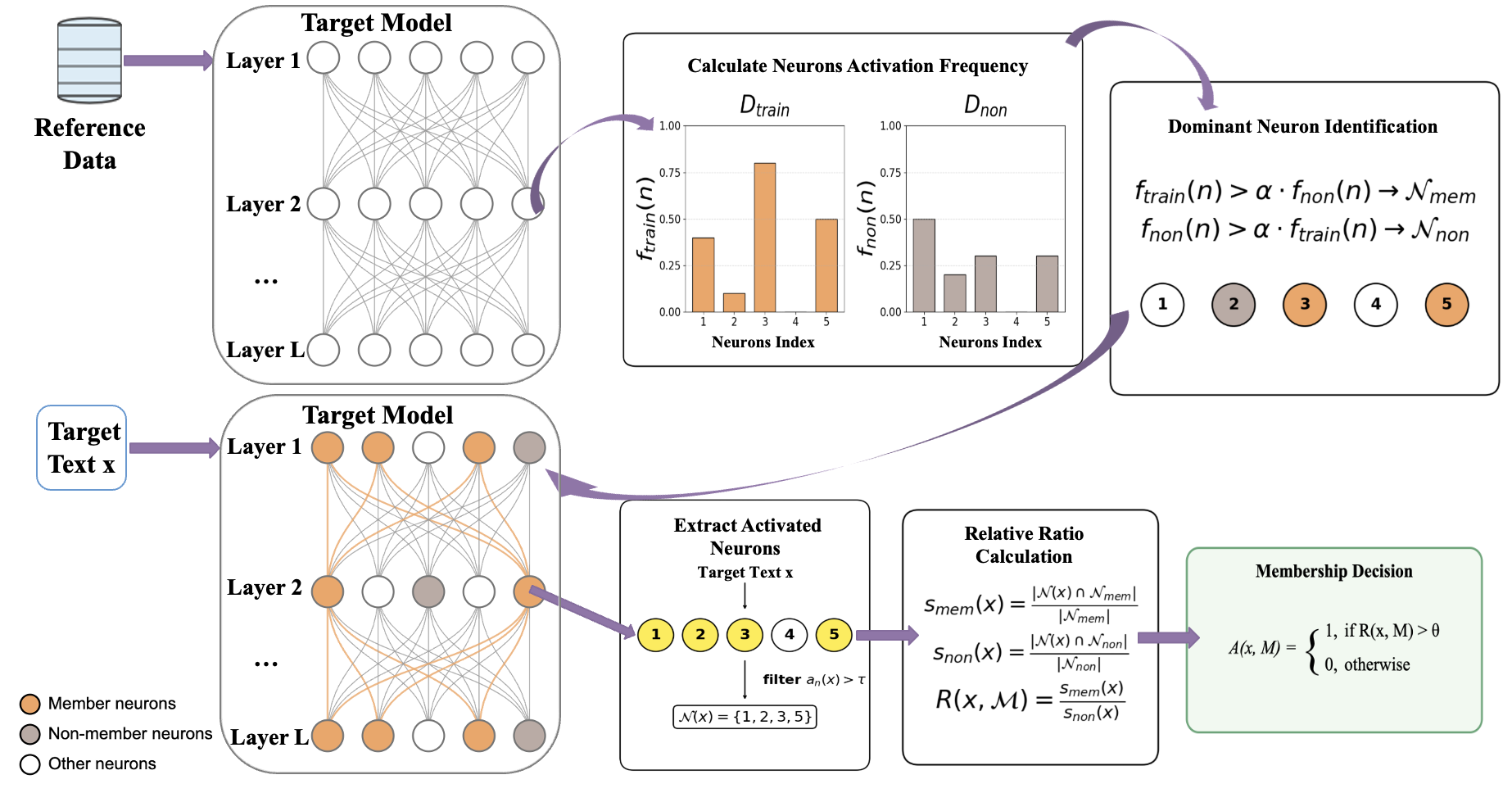}
  \caption{An overview of NA-PDD.}
  \label{fig:methodological-procedure}
\end{figure*}

\section{Related Work}
\noindent \textbf{Membership Inference Attacks (MIA).} Membership inference attacks determine whether specific data was used to train a model~\cite{shokri2017membership, hu2022membership}. Originating in genomics~\cite{homer2008resolving, pyrgelis2017knock}, this field evolved within machine learning through shadow modeling and black-box techniques~\cite{salem2018ml, yeom2018privacy}. MIA research has expanded across computer vision~\cite{choquette2021label}, generative models~\cite{chen2020gan}, and diffusion systems~\cite{carlini2023extracting}, while defensive strategies like differentially private training emerged in parallel~\cite{abadi2016deep, jia2019memguard}. Recently, MIA has become crucial for Large Language Models, detecting memorized training data~\cite{nasr2023scalable, oren2023proving} and potential copyright issues~\cite{duarte2024cop, meeus2024did}. Our work focuses specifically on the the precise detection of pre-training data.

\noindent\textbf{LLM Pretraining Data Detection.} Traditional MIAs predominantly employ black-box approaches, relying solely on model output signals for inference~\cite{yeom2018privacy, sablayrolles2019white}. In contrast, we adopt a white-box strategy, directly accessing the model's internal states. In NLP, previous research includes likelihood ratio attacks on causal language models by \citet{carlini2021extracting}, neighborhood attacks proposed by \citet{mattern2023membership}, and membership inference based on outlier word likelihoods and probability distribution features by \citet{shi2023detecting}, \citet{mireshghallah2022quantifying}, and \citet{watson2021importance}. Unlike these approaches that depend on surface features, we investigate LLM internal neuron activation states, exploring their specific memorization characteristics for pre-training data, thereby achieving enhanced detection performance.

\section{Methodology}
In this section, we begin by introducing the task of pre-training data detection in large language models. We then provide a detailed description of our method, NA-PDD, a pre-training data detection algorithm that captures the differences in neuronal activation between the pre-training corpus and other data.

\subsection{Problem Statement}
Pre-training data detection (also known as membership inference) aims to determine whether a large language model (LLM) has utilized specific data points, such as text, during its training phase. Formally, for a given text $x$ and a target LLM $\mathcal{M}$, the detection algorithm is structured as a binary classification problem:
\begin{equation}
\mathcal{A}(x, \mathcal{M}) \rightarrow \{0, 1\}
\end{equation}
where a prediction of \textbf{1} indicates that the model M has utilized the text $x$, while a prediction of \textbf{0} signifies that it has not.

\textbf{White-box settings}. Following previous work, we assume access to the weights and activations of the target model. This framework is applicable in two real-world scenarios: 1) Model owners need to audit their models to prevent dataset contamination \cite{xu2024benchmark, magar2022data} or to assess the effectiveness of machine unlearning \cite{bourtoule2021machine, yao2024machine}; 2) Data owners need to verify whether their data has been incorporated into an open-source model \cite{mokander2024auditing, pan2020privacy}.

\subsection{Overview}
We now propose NA-PDD, which is outlined in Figure \ref{fig:methodological-procedure}. Our design is based on the observation that the language model tends to activate specific neurons when inferring with training data as opposed to other data. NA-PDD consists of the following four steps: (i) \textit{Neuronal Activation Determination}: This phase detects active neurons by analyzing their activation values as the target model $\mathcal{M}$ processes the given text $x$ (Section \ref{step1}). (ii) \textit{Neuronal Identity Discrimination}: By examining the activation status of neurons in response to different types of data (training data and non-training data), we label neurons that are more easily activated by training data as "member neurons," while those that respond more to non-training data are labeled as "non-member neurons" (Section \ref{step2}). (iii) \textit{Neuronal Similarity Calculation}: To ascertain whether a target text $x$ was part of the training data for a large language model, we evaluate the relationship between member neurons and $x$ by examining the neurons activated when $x$ is input into $\mathcal{M}$ (Section \ref{step3}). (iv) \textit{Membership Inference}: By comparing the advantage of member neurons for the target text $x$ across different layers against a predefined threshold, we predict whether the text was part of the pre-training corpus of the target model (Section \ref{step4}). Our method is summarized in Algorithm \ref{alg:algorithm}, which is detailed in Appendix~\ref{app:Algorithm}.

\subsection{Neuronal Activation Determination}
\label{step1}
The sparsely activated nature of large language models suggests \cite{wang2024q, liu2024training} that only a few neurons are activated when processing text $x$. These neurons have a significant influence on the prediction of $x$. To identify which neurons play a major role in the inference process of the target model on specific training corpora (e.g., certain news texts used for pre-training), we establish a flexible threshold $\tau$ to determine the activation state of neurons. 

Formally, given input text $x$ to the target model $\mathcal{M}$, which consists of neurons $\mathcal{N}$, the activation state of neuron $n \in \mathcal{N}$ is defined as follows:
\begin{equation}
I(x,n) =
\begin{cases}
1, & \text{if } a_n(x) > \tau,\\
0, & \text{otherwise}.
\end{cases}
\label{activate_rule}
\end{equation}

where $a_n(x)$ denotes the output value of neuron $n$ for text $x$. Neurons with an output value greater than the threshold $\tau$ are considered to be "active".

\subsection{Neuronal Identity Discrimination}
\label{step2}
We introduce reference corpora to identify neurons that tend to be activated when processing training or non-training data. The reference data can originate from the model owner's training database \cite{liu2024datasets}. Specifically, for a set of collected training data $\mathcal{D}_{train}$, we record the activation frequency of all neurons $n$ in the target model $\mathcal{M}$:
\begin{equation}
\label{eq:freq_mem}
f_{train}(n) = \frac{1}{|\mathcal{D}_{train}|} \sum_{x \in \mathcal{D}_{train}} I(x, n)
\end{equation}

where $f_{train}(n)$ represents the frequency at which neurons $n$ are activated for the training data. The frequency of neuron activation for non-training data is calculated in a similar manner:
\begin{equation}
\label{eq:freq_non}
f_{non}(n) = \frac{1}{|\mathcal{D}_{non}|} \sum_{x \in \mathcal{D}_{non}} I(x, n)
\end{equation}

The frequency at which neurons are activated can help identify which neurons play a more significant role in the prediction process. However, relying solely on these activation frequencies does not effectively distinguish between the activation of neurons for training texts and non-training texts. For some low-level lexical neurons, activation occurs frequently for both types of texts as these neurons extract common features. Consequently, focusing on these neurons is insufficient for confirming whether the model is utilizing specific data during training. Therefore, we assess the relative dominance of neuron activation frequency to determine whether a neuron is dominant in the training data, using the following formula:
\begin{equation}
\mathcal{N}_{mem} = \{n \in \mathcal{N} \mid f_{train}(n) > \alpha \cdot f_{non}(n) \}
\label{eq:member_dominant}
\end{equation}

Neurons that are more likely to activate in response to training texts, compared to non-training texts, are referred to as "member" neurons. Similarly, the computation process for identifying "non-member" neurons is as follows:
\begin{equation}
\mathcal{N}_{non} = \{n \in \mathcal{N} \mid f_{non}(n) > \alpha \cdot f_{train}(n) \}
\label{eq:nonmember_dominant}
\end{equation}

where $\alpha > 1$ serves as the dominance threshold, controlling the stringency for classifying neurons as "member" or "non-member" neurons.

\subsection{Neuronal Similarity Calculation}
\label{step3}
The powerful learning capabilities of LLMs often lead to noticeable memory effects on training data. This characteristic is a key reason why pre-training data detection algorithms are effective \cite{carlini2021extracting, tirumala2022memorization}. We demonstrate that, beyond mere superficial features or hidden units, the activation patterns of neurons can more accurately reflect the memory phenomena within language models. Specifically, the model’s retention and assimilation of the training corpus enable it to pinpoint specific neurons that enhance and accelerate the reasoning process for similar texts. Building on this intuition, if text $x$ was used in training the target model $\mathcal{M}$, the neurons activated during the inference process will show a higher similarity to those activated by other training data. We measure this similarity through the coincidence rate of activated neurons.
\begin{equation}
\label{eq:ratio_mem}
s_{mem}(x) = \frac{|\mathcal{N}(x) \cap \mathcal{N}_{mem}|}{|\mathcal{N}_{mem}|}
\end{equation}

where $\mathcal{N}(x)$ represents the set of neurons activated for text $x$ according to Eq.\ref{activate_rule}. Similarly, the coincidence rate of neurons activated by the text $x$ with non-member neurons is calculated as follows:
\begin{equation}
\label{eq:ratio_non}
s_{non}(x) = \frac{|\mathcal{N}(x) \cap \mathcal{N}_{non}|}{|\mathcal{N}_{non}|}
\end{equation}

\subsection{Membership Inference}
\label{step4}
Different layers of large language models often undertake diverse roles during the inference process \cite{zhao2024explainability} and exhibit varying performance on pre-training data detection task \cite{liu2024probing}. Therefore, we propose a scoring mechanism to evaluate the capability of different neuronal layers to differentiate between training and non-training data as follows:
\begin{equation}
 S_\ell = |\mathcal{N}^\ell_{mem}| - |\mathcal{N}^\ell_{non}|
\label{eq:discriminative_score}
\end{equation}

where $\mathcal{N}^\ell_{mem}$ and $\mathcal{N}^\ell_{non}$ represent the member and non-member neurons of the $\ell$-layer, respectively. The higher the score, the greater the imbalance between member and non-member neurons within that layer.
Based on the difference scores $S_\ell$, we select the $K$ most discriminative layers $\mathcal{L}_{dis}$ and compute the average activation neuron similarity from the selected layers:

\begin{equation}
\label{eq:agg_mem}
\bar{s}_{mem}(x) = \frac{1}{|\mathcal{L}_{dis}|} \sum_{\ell \in \mathcal{L}_{dis}} s^\ell_{mem}(x)
\end{equation}

\begin{equation}
\label{eq:agg_non}
\bar{s}_{non}(x) = \frac{1}{|\mathcal{L}_{dis}|} \sum_{\ell \in \mathcal{L}_{dis}} s^\ell_{non}(x)
\end{equation}

The member advantage for text $x$ is defined by the ratio of $s_{mem}(x)$ to $s_{non}(x)$. A higher ratio suggests a greater likelihood that text $x$ was part of the training data for the target model $\mathcal{M}$.
\begin{equation}
\label{eq:membership_score}
R(x, \mathcal{M}) = \frac{\bar{s}_{mem}(x)}{\bar{s}_{non}(x)}
\end{equation}

After calculating the member advantage $R(x, \mathcal{M})$ for text $x$, we predict whether $x$ was included in model $\mathcal{M}$'s pretraining data by applying a predefined threshold $\theta$ to $R(x, \mathcal{M})$:
\begin{equation}
\mathcal{A}(x, \mathcal{M}) =
\begin{cases}
1, & \text{if } R(x, \mathcal{M}) > \theta,\\
0, & \text{otherwise}.
\end{cases}
\label{activate_result}
\end{equation}

\section{Data Construction}
As proprietary information, the training logs of LLMs are generally not publicly accessible. To evaluate the performance of pre-training data detection (PDD) algorithms on LLMs, researchers typically rely on commonly used benchmark datasets (e.g., Wikipedia) as proxies for the models' training data. Non-training data, in contrast, is curated based on the release timelines of the target LLMs \cite{shidetecting, liu2024probing}. For example, the Pythia model (2023 release) \cite{biderman2023pythia} cannot have been trained on Wikipedia articles published after 2024. However, this temporal partitioning of training versus non-training data introduces a critical time drift confounder. This methodological limitation raises questions about whether the observed efficacy of PDD algorithms genuinely reflects their discriminative capability or merely artifacts of the dataset construction methodology (e.g., temporal distributional shifts rather than true memorization signals) \cite{maini2025reassessing, duanmembership}.

To address this limitation, we propose \textbf{CCNewsPDD}, a carefully designed benchmark for PDD evaluation that effectively eliminates temporal confounding effects. Our approach utilizes the CCNews corpus \cite{Hamborg2017}, which is known to have been part of the training data for major LLMs including Pythia \cite{biderman2023pythia} and OPT \cite{zhang2022opt}. We specifically selected news articles published in August 2017 to ensure complete temporal consistency across all evaluated data.

To generate novel text that maintains authentic linguistic properties while ensuring that the target LLMs have not encountered it during training, we engage in a data transformation process. We begin by selecting half of the CCNews corpus, followed by the implementation of three principled data transformations. These transformations are meticulously designed to preserve the original data distribution while introducing meaningful variations, thereby creating a rigorous testbed for evaluating the PDD algorithm. From this process, we derive three complementary datasets that together offer a comprehensive assessment framework free from temporal bias:

\textbf{CCNewsPDD(trans):} This dataset implements sequential translation from English to French and back to English using MarianMT models \cite{tiedemann2024democratizing}. This approach generates semantically preserved non-training data with diverse syntactic variations. % \cite{edunov-etal-2018-understanding}

\textbf{CCNewsPDD(mask):} This dataset randomly masks 15\% of tokens in the original text and uses BERT \cite{devlin2019bert} to predict contextually appropriate substitutions. Unlike back-translation, this strategy focuses on localized perturbations while maintaining the global text structure.

\textbf{CCNewsPDD(prompt):} This dataset generates non-training data through explicit instruction prompting. By directing the BART model \cite{lewis-etal-2020-bart} to reformulate the original texts, we achieve comprehensive discourse-level rephrasings while preserving the core semantic content in the generated non-training data.

\section{Experimental Settings}

\begin{table*}[t]
\centering
\small
\renewcommand{\arraystretch}{1.2}
\begin{tabular}{lllll}
\toprule
\textbf{Benchmark} & \textbf{Data source} & \textbf{Text length} & \textbf{\#Examples} & \textbf{Applicable models} \\
\midrule
WikiMIA \cite{shi2023detecting} & Wikipedia & 32 & 774 & Pythia-2.8B, OPT-6.7B \\
ArxivMIA \cite{liu2024probing} & Arxiv & 143.1 & 2,000 & TinyLLaMA-1.1B, OpenLLaMA-13B \\
CCNewsMIA (\textbf{Ours}) & CC-news & 309.1 & 1,200 & Pythia-2.8B, OPT-6.7B \\
\bottomrule
\end{tabular}
\caption{Benchmark summary statistics: Each benchmark has an equal split of training and non-training examples. "Text Length" refers to the average number of tokens in each text example of the benchmark. "\#Examples" denotes the total number of text examples in the benchmark.}
\label{tab:benchmarks}
\end{table*}

\textbf{Benchmarks and Models.} We evaluate NA-PDD's performance across three benchmark datasets (Table \ref{tab:benchmarks}). For ArxivMIA \cite{liu2024probing}, comprising arXiv academic paper abstracts, we follow the original study's experimental setting by testing TinyLLaMA-1.1B \cite{zhang2024tinyllama} and OpenLLaMA-13B \cite{geng2023openllama}. For WikiMIA \cite{shi2023detecting} and our CCNewsPDD benchmark, we select Pythia-2.8B \cite{biderman2023pythia} and OPT-6.7B \cite{zhang2022opt} as our evaluation models, as their pretraining corpora are known to include Wikipedia dumps and the CCNews dataset, respectively. 

\noindent \textbf{Baselines.} We evaluate NA-PDD against nine state-of-the-art pretraining data detection (PDD) methods, which include both reference-free and reference-based approaches. The reference-free methods are: (i) \textit{Loss Attack} \cite{yeom2018privacy}, which utilizes the target model's loss values; (ii) \textit{Neighbor Attack} \cite{mattern2023membership}, which compares loss values between target samples and synthetically generated neighbor texts; (iii) \textit{Min-K\% Prob} \cite{shi2023detecting}, which analyzes the average log-likelihood of the least probable tokens; (iv) \textit{Min-K\%++ Prob} \cite{zhang2024min}, which extends this approach with vocabulary-normalized probabilities; and (v) \textit{DC-PDD} \cite{zhang2024pretraining}, which proposes a divergence-based calibration Method for pretraining data detection. 

The reference-based methods, which employ a reference (proxy) model for detection calibration, include: (vi) \textit{Zlib} \cite{carlini2021extracting}, which computes the ratio between an example's perplexity and its zlib entropy; (vii) \textit{Lowercase} \cite{carlini2021extracting}, which compares perplexity between original and lowercased text; (viii) \textit{Small Ref} \cite{carlini2021extracting}, which uses perplexity ratios between target and reference models; and (ix) \textit{Probe Attack} \cite{liu2024probing}, which examines internal model activations through probing techniques. 

\noindent \textbf{Evaluation Metrics.} Following prior work \cite{shi2023detecting,zhang2024pretraining}, we use the Area Under the ROC Curve (AUC) to assess detection performance. AUC provides a threshold-independent measure of pretraining data detection performance and is robust to class imbalance.

\noindent \textbf{Implementation Details.} We introduce the details of NA-PDD through three core components: (i) \textit{Neuronal Activation Determination}, where we attach hook functions to all feed-forward network (FFN) layers in the Transformer architecture to record post-activation outputs of neurons; (ii) \textit{Neuronal Identity Discrimination}, which utilizes 100 training and 100 non-training samples to classify neurons as either member or non-member neurons; and (iii) \textit{Hyperparameter Selection}, where extensive ablation studies determine the final configuration: activation threshold $\tau = 1.0$ (validated range [0,2]), dominance threshold $\alpha = 1.5$ (tested range [1.2,2.0]), and the number of used discriminative layers $K=10$ (tested range [1, 32]). For more details on our baselines, please refer to Appendix~\ref{app:implementation}.

%For more details on these implementation details, please refer to Appendix~\ref{app:implementation}.

\section{Experimental Results}
We report the results of the experiment to address the following questions: \textbf{Q1}: What is the performance of NA-PDD across different datasets and language models? \textbf{Q2}: How do the size of the model and the amount of reference data impact the performance of the PDD algorithm? \textbf{Q3}: Sensitivity analysis: How does NA-PDD perform at different activation threshold $\tau$, dominance threshold $\alpha$, and numbers of used discriminative layers $K$?

\subsection{Main Results}
Table \ref{tab:results} presents the performance comparison of the PDD algorithm across different datasets, leading to four key findings: 

\noindent \textbf{Leading Performance of NA-PDD}: Our method \\ achieves the best performance on all datasets. Notably, in detecting whether the CCNewsPDD(prompt) dataset was used in OPT model training, NA-PDD significantly increases the AUROC value from 90.1\% to 99.7\% compared to the second best algorithm. This confirms neuronal activation's utility in PDD, as NA-PDD reliably distinguishes training data through activation pattern analysis.

\noindent \textbf{Limited Advantages of Reference Model-Based Approaches}:  While reference-based methods necessitate extra computational resources for prediction calibration, their performance gains are not significant. For example, when evaluating OPT in CCNewsPDD(trans), the leading reference-based method, Probe Attack (68.8\% AUROC), was outperformed by the reference-free method, Min-K++\% Prob (76.6\% AUROC). This indicates that reference-free methods retain potential, as evidenced by our NA-PDD approach, which achieves outstanding detection results solely through neuronal activation analysis.

\noindent \textbf{Detection Difficulty Variations Between Datasets}: The ArxivMIA dataset posed significant detection challenges, with the highest AUC (57.2\%) among all methods being considerably lower than those for WikiMIA (75.8\%) and CCNewsPDD (99.7\%). We speculate that this may be due to the technical nature of ArxivMIA, which makes it more challenging for models to memorize training texts and thus complicates PDD task.

\noindent \textbf{Detection Difficulty Variations Between Models}: On the CCNewsPDD(mask) dataset, NA-PDD achieved better performance with the smaller Pythia-2.8B model (98.8\% AUC) compared to the larger OPT-6.7B model (95.2\% AUC). This performance gap likely arises from their differing training paradigms: Pythia employs pure causal language modeling, whereas OPT uses standard autoregressive language modeling. This highlights the importance of developing PDD algorithms that are specifically tailored to different training modes.

\begin{table*}[t]
\centering
\resizebox{\textwidth}{!}{%
\begin{tabular}{lccccccccccc}
\hline
\multirow{2}{*}{\textbf{Method}} & \multicolumn{2}{c}{\textbf{ArxivMIA}} & \multicolumn{2}{c}{\textbf{WikiMIA}} & \multicolumn{2}{c}{\textbf{CCNewsPDD(trans)}} & \multicolumn{2}{c}{\textbf{CCNewsPDD(mask)}} & \multicolumn{2}{c}{\textbf{CCNewsPDD(prompt)}} \\
& \textbf{TinyL.} & \textbf{OpenL.} & \textbf{Pythia} & \textbf{OPT} & \textbf{Pythia} & \textbf{OPT} & \textbf{Pythia} & \textbf{OPT} & \textbf{Pythia} & \textbf{OPT} \\
\hline
\rowcolor{gray!20}
\multicolumn{11}{l}{\textbf{Reference-free}} \\
Loss Attack & 45.1 & 49.1 & 65.8 & 64.5 & 55.2 & 52.6 & 75.7 & 68.5 & 71.0 & 70.3 \\
Neighbor Attack & 55.9 & 56.2 & 66.0 & 65.6 & 63.8 & 61.7 & 78.1 & 77.0 & 44.4 & 46.2 \\
Min-K\% Prob & 45.5 & 49.2 & 64.2 & 64.9 & 60.6 & 62.4 & 76.8 & 72.7 & 70.5 & 71.8 \\
Min-K++\% Prob & 47.8 & 52.0 & 64.9 & 67.3 & 71.0 & 76.6 & 68.7 & 74.2 & 66.9 & 73.0 \\
DC-PDD & 46.2 & 48.7 & 63.3 & 65.4 & 59.4 & 60.7 & 76.5 & 72.1 & 72.1 & 71.8 \\
\hline
\rowcolor{gray!20}
\multicolumn{11}{l}{\textbf{Reference-based}} \\
Zlib Compression & 43.0 & 43.8 & 66.5 & 65.1 & 69.9 & 68.0 & 79.4 & 75.6 & 81.7 & 80.7 \\
Lowercased Text & 45.8 & 48.8 & 60.9 & 61.5 & 57.5 & 55.6 & 64.7 & 62.4 & 72.8 & 74.7 \\
Smaller Model & {---} & 56.7 & 58.1 & 63.4 & 54.0 & 66.2 & 48.1 & 55.9 & 40.6 & 60.9 \\
Probe Attack & 53.2 & 59.0 & 69.8 & 68.1 & 65.3 & 68.8 & 81.3 & 82.3 & 90.7 & 90.1 \\
\hline
\rowcolor{gray!20}
\multicolumn{11}{l}{\textbf{Our Method}} \\
NA-PDD & \textbf{57.2} & \textbf{59.3} & \textbf{75.8} & \textbf{71.6} & \textbf{92.4} & \textbf{92.1} & \textbf{98.8} & \textbf{95.2} & \textbf{99.6} & \textbf{99.7} \\
\hline
\end{tabular}%
}
\caption{AUC scores (in \%) for various methods across ArxivMIA, WikiMIA, and CCNewsPDD datasets. Best performances in each column are highlighted in bold.}
\label{tab:results}
\end{table*}

\subsection{Impact of Different Factors}
In this section, we analyze how model size and the amount of reference data affect the performance of the PDD algorithm. To highlight the effectiveness and robustness of NA-PDD, we compare it with Probe Attack, the second-best algorithm according to our main results.

\begin{figure}
    \centering
    \includegraphics[width=0.8\linewidth]{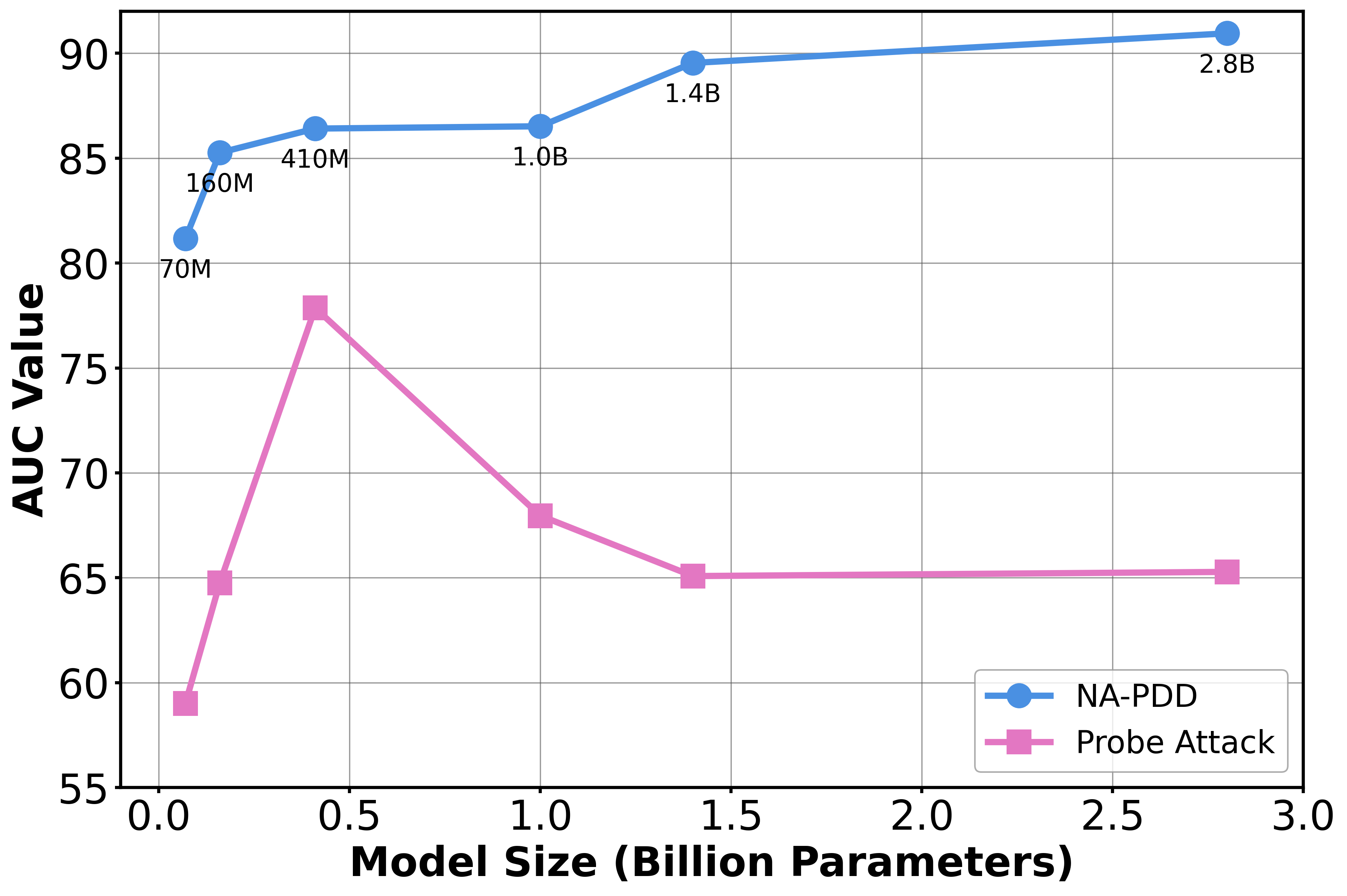}
    \caption{Comparison of AUC Values with Different Model Sizes (best viewed in color).}
    \label{fig:model_size}
\end{figure}

\textbf{Model Size.} We compare NA-PDD with the Probe Attack baseline across various sizes of Pythia models (ranging from 70M to 2.8B) on the CCNewsPDD(trans) dataset. As shown in Figure~\ref{fig:model_size}, our method consistently improves with increasing model size, indicating that larger models develop more distinctive neuron activation patterns for detecting pretraining data. In contrast, the baseline method exhibits unstable behavior: after peaking at 78\% AUC with the 410M model, its performance sharply declines and stabilizes around 65\% for models larger than 1.4B. This divergence suggests that while traditional probe-based approaches struggle with increased model complexity, our neuron-based detection method effectively utilizes the richer representations in larger models. The widening performance gap highlights the scalability advantage of our approach.

\begin{figure}
    \centering
    \includegraphics[width=0.8\linewidth]{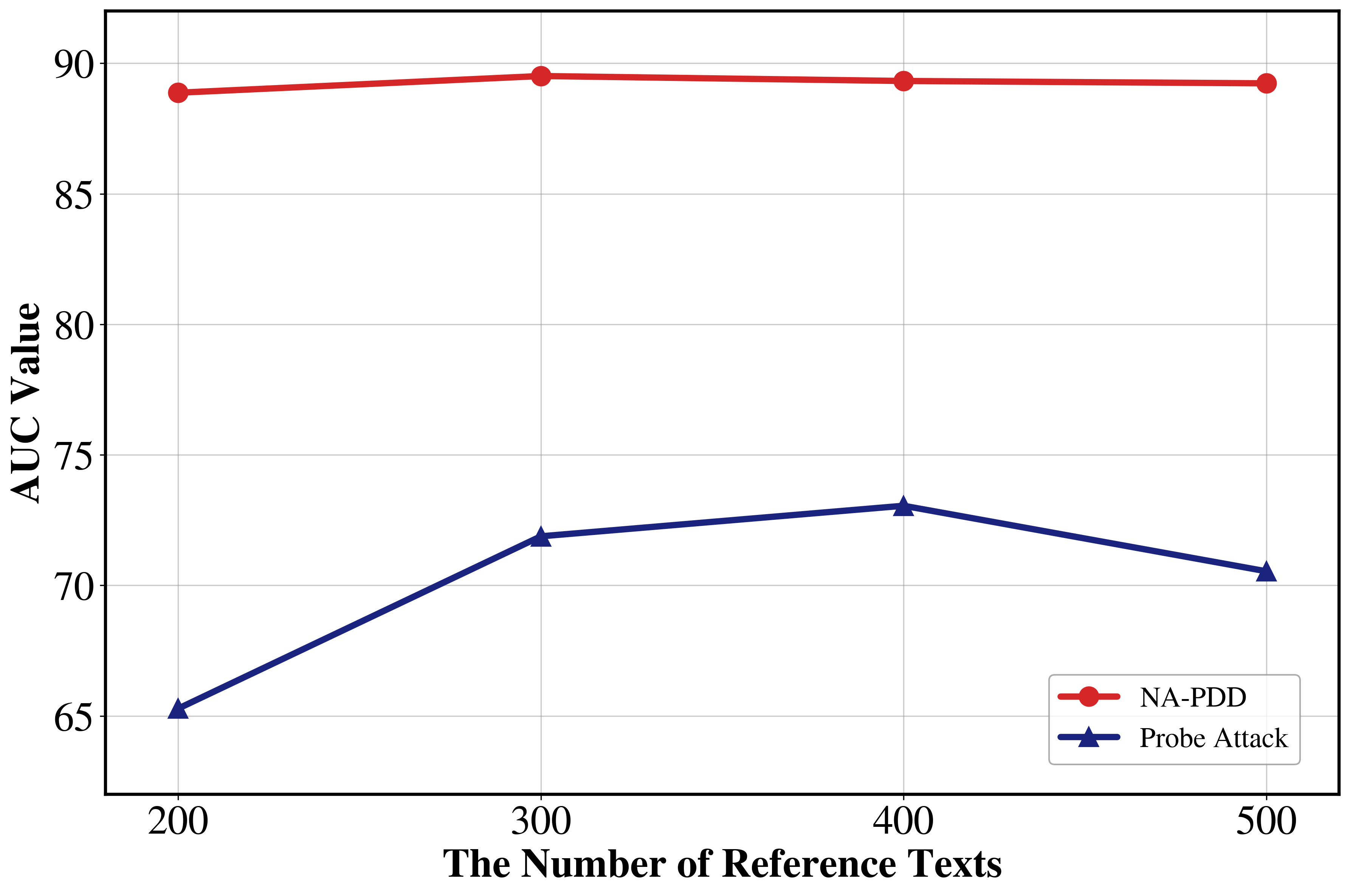}
    \caption{Comparison of AUC Values with Different Training Data Sizes (best viewed in color).}
    \label{fig:data_size}
\end{figure}

\textbf{Reference Data Size.} We assess the robustness of our method by varying the reference data sizes (200-500 samples) for neuron classification using the Pythia-2.8B model on the CCNewsPDD(trans) dataset. As illustrated in Figure~\ref{fig:data_size}, our method consistently achieves high performance (nearly 90\% AUC) across all data sizes, highlighting its data efficiency. In contrast, the baseline Probe Attack is more sensitive to the volume of reference data. Our approach maintains high performance with limited reference data and significantly outperforms all baselines by 16--34\%, demonstrating both its data efficiency and robustness.

\subsection{Sensitivity Analysis}
\label{sec:Sensitivity Analysis}
We further conduct sensitivity analyses on three key hyperparameters: the activation threshold $\tau$, the dominance threshold $\alpha$, and the numbers of selected discriminative layers $K$.

\vspace{0.3em} 

\begin{figure}
    \centering
    \includegraphics[width=0.9\linewidth]{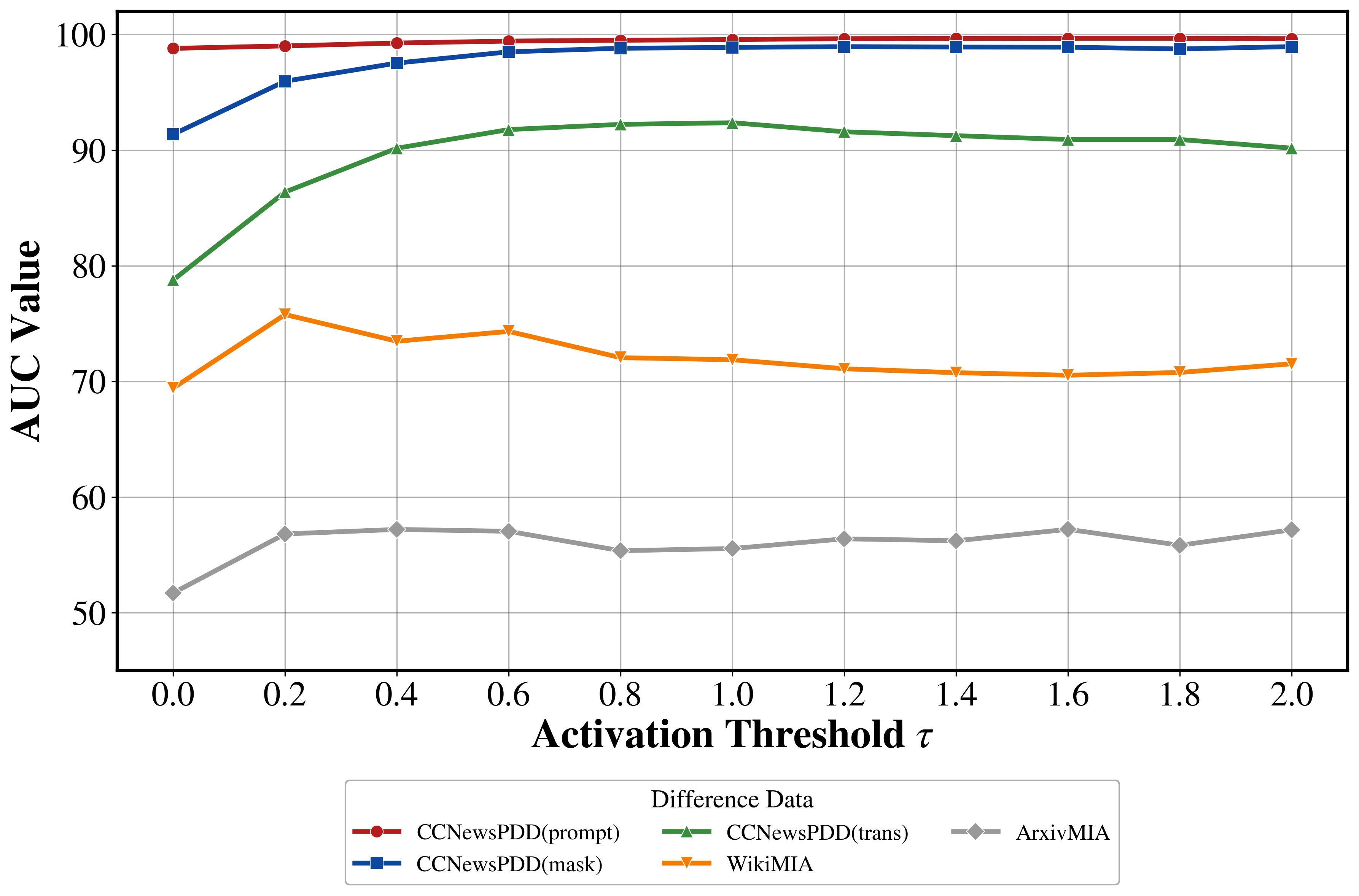}
    \caption{Comparison of AUC Values with Different Activation Thresholds.}
    \label{fig:activation_threshold}
\end{figure}

\textbf{Activation Threshold $\tau$.} We evaluate the robustness of our membership inference approach with varying activation thresholds $\tau \in [0.0, 2.0]$ across three distinct benchmarks. As illustrated in Figure~\ref{fig:activation_threshold}, our method shows remarkable resilience to changes in activation threshold, despite performance differences across datasets. For instance, CCNewsPDD(prompt) and CCNewsPDD(mask) variants consistently reach peak performance with an AUC of approximately 99\% across the entire threshold range. In contrast, the CCNewsPDD(trans) variant, despite starting at a lower performance, stabilizes above 90\% for $\tau \geq 0.2$. 

%The WikiMIA dataset shows only minor performance fluctuations, with a variation of just 6\% (70\%–76\%), while ArxivMIA demonstrates exceptional stability with less than a 5\% deviation. These findings emphasize the inherent robustness of our approach to hyperparameter selection, ensuring reliable performance across diverse data distributions without the need for extensive threshold tuning.

\vspace{0.3em}  

\begin{figure}
    \centering
    \includegraphics[width=0.8\linewidth]{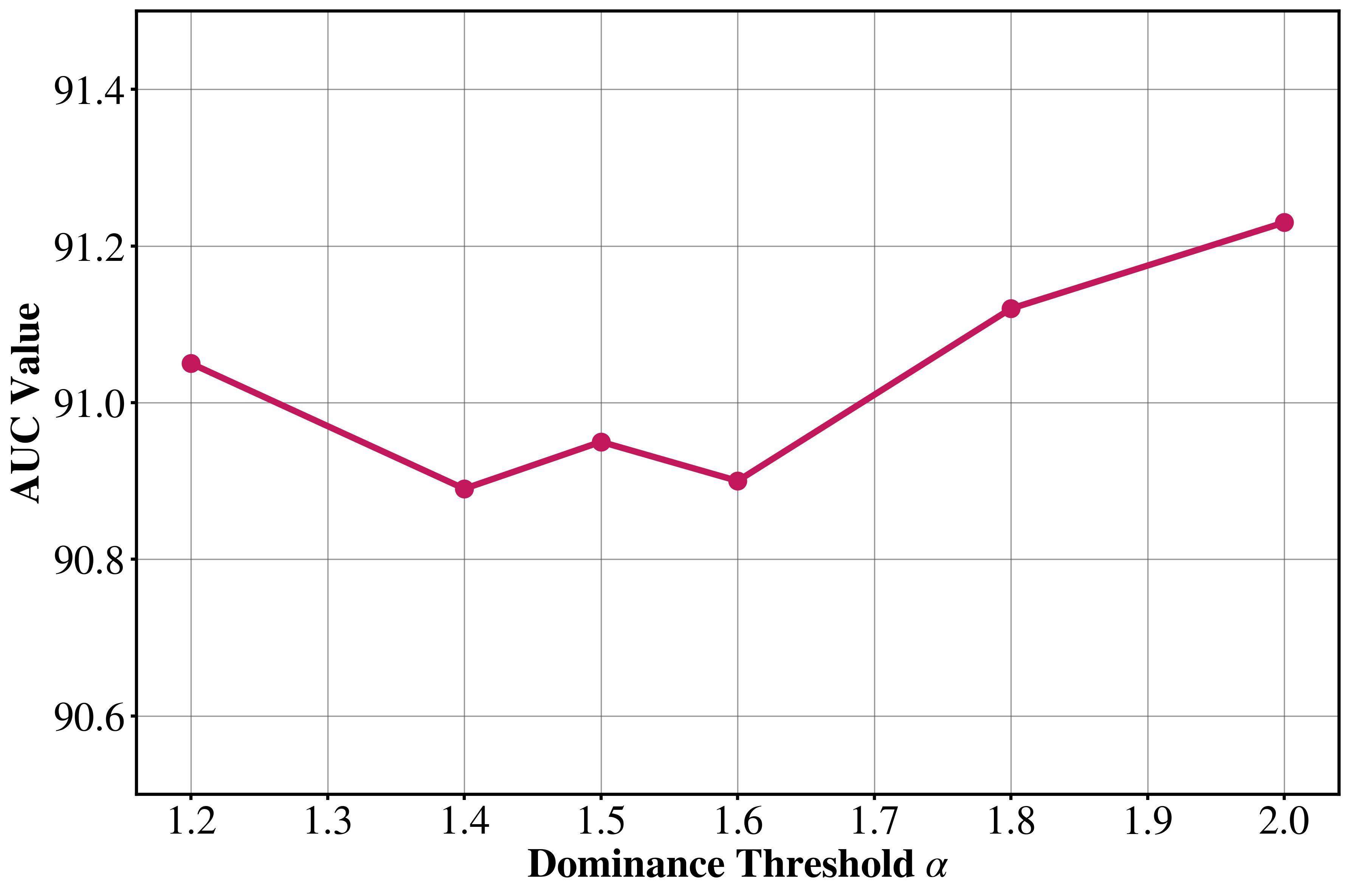}
    \caption{Comparison of AUC Values with Different Dominance Thresholds.}
    \label{fig:Alpha_threshold}
\end{figure}

\textbf{Dominance Threshold $\alpha$.} We assess the impact of the dominance threshold $\alpha$ on the CCNewsPDD(trans) dataset. As depicted in Figure \ref{fig:Alpha_threshold}, the AUC remains stable (90.9\%–91.25\%) across different $\alpha$ values, exhibiting a slight U-shaped trend: starting at 91.05\% ($\alpha = 1.2$), dipping to 90.90\% ($\alpha = 1.4$), and then peaking at 91.25\% ($\alpha = 2.0$). The minimal variation of 0.35\% demonstrates the robustness of our method to the selection of $\alpha$, with $\alpha = 1.5$ recommended as a reliable default.

\vspace{0.3em}  

\begin{figure}
    \centering
    \includegraphics[width=0.8\linewidth]{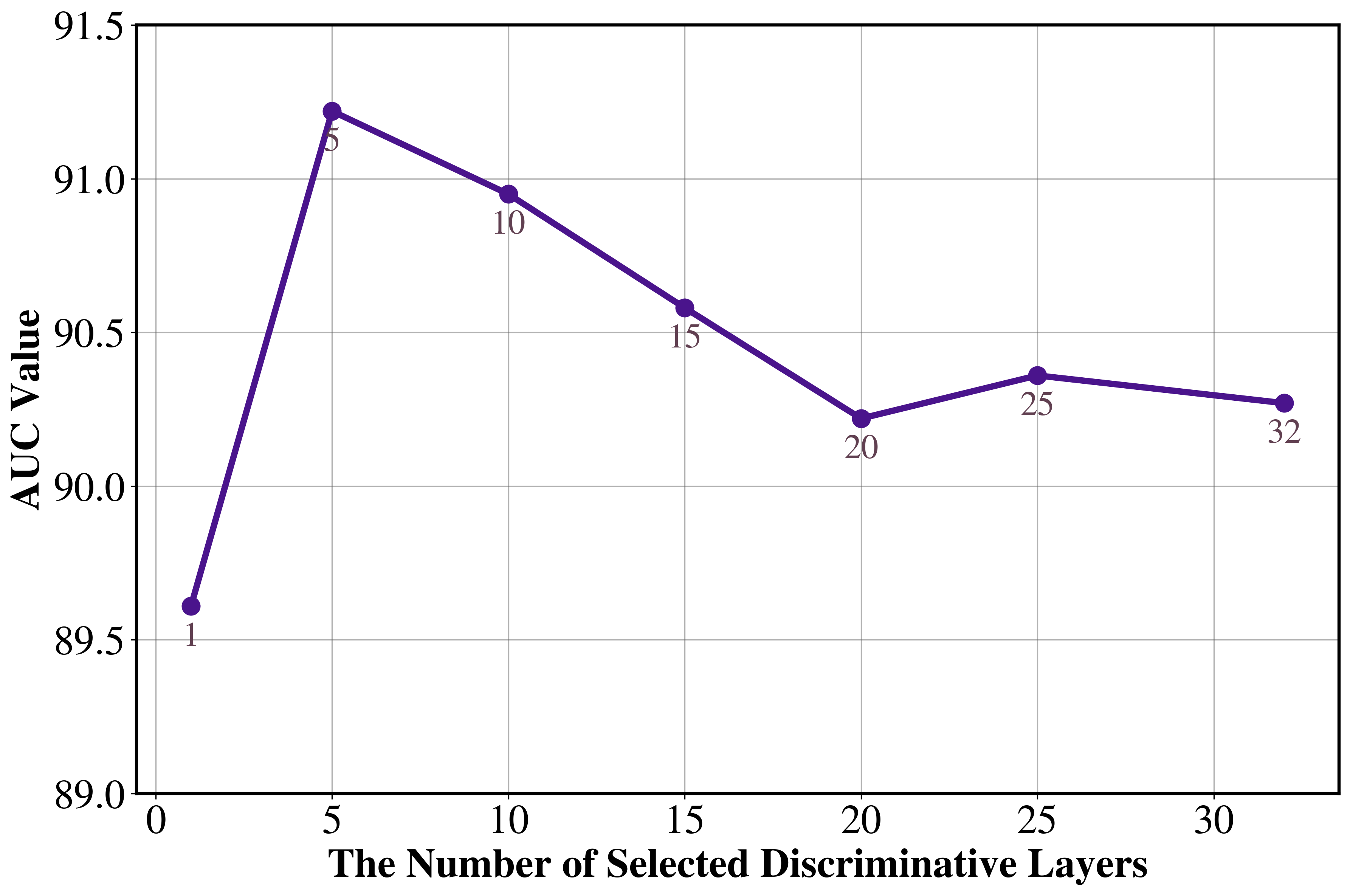}
    \caption{Comparison of AUC Values with Different Numbers of Selected Discriminative Layers.}
    \label{fig:top-k_layers}
\end{figure}

\textbf{Numbers of selected discriminative layers $K$.} Our evaluation on the Pythia-2.8B model and CCNewsPDD(trans) dataset shows stable detection performance of NA-PDD across varying $K$ values (Fig.~\ref{fig:top-k_layers}): AUC increases from 89.6\% ($K=1$) to a peak of 91.2\% ($K=5$), maintaining over 90\% for $K \geq 5$ (range: 90.2\%–91.2\%). The minimal variation highlights our algorithm's robustness, with $K=5$ providing a good balance between efficiency and performance, while permitting flexible selection of $K$ (from 5 to 32) without significant performance loss.

\section{Conclusion}
In this paper, we propose NA-PDD, a PDD algorithm that distinguishes between training and non-training data by analyzing neuronal activation during LLM inference. Our experimental results show that NA-PDD outperforms baselines across various LLMs and evaluation benchmarks, demonstrating robust performance and insensitivity to hyperparameters. In future work, we aim to explore more complex neuronal activation patterns, such as how activation pathways can further enhance PDD. \\

\noindent \textbf{Limitations}

\noindent While NA-PDD shows promising results in detecting pretraining data, it has several limitations. (i) NA-PDD relies on capturing neuron activation patterns for both training and non-training data. Consequently, it is only applicable when the weights of large language models (LLMs) and activation information are accessible. This limitation means it cannot be used with closed-source models or LLMs that only provide a query interface. (ii) NA-PDD requires a portion of a reference corpus to help label member and non-member neurons. Although obtaining these reference corpora is feasible in scenarios like copyright verification and LLM internal audits, and the amount needed is relatively small, NA-PDD is not suitable for situations lacking a reference corpus. (iii) Due to computational resource constraints, NA-PDD has only been evaluated on LLMs with up to 13 billion parameters. However, previous work and our experiments suggest that the PDD algorithm performs better with larger LLMs, making the potential performance of NA-PDD on larger models promising. We plan to explore this in future work.

\bibliography{zhihao}

%\clearpage
\appendix
\section{Algorithm}
\label{app:Algorithm}

\begin{algorithm}[H]     
  \caption{Neuron Activation Pre-training Data Detection (NAPDD)}
  \label{alg:algorithm}
  \begin{algorithmic}[1]
\Require
  $x$; $M$ with neurons $\mathcal N$, $L$ layers;\\
  $D_{\mathrm{train}},D_{\mathrm{non}}$; $\tau,\alpha>1.5,\theta$; $K\le L$
\Ensure
  $A(x,M)\in\{0,1\}$

\State Forward $x$ through $M$; collect $\{a_n(x)\}$.

\ForAll{$n\in\mathcal N$}
  \State Extract activation $I(x,n)=1[a_n(x)>\tau]$, {w.r.t.\ Eq.~\ref{activate_rule}}
\EndFor

\ForAll{$n\in\mathcal N$}
  \State Compute activation frequency $f_{\mathrm{train}}(n)$ and $f_{\mathrm{non}}(n)$, {w.r.t.\ Eqs.~\ref{eq:freq_mem},\,\ref{eq:freq_non}}
\EndFor

\State Define  dominant neuron $N_{\mathrm{mem}},N_{\mathrm{non}}$, {w.r.t.\ Eqs.~\ref{eq:member_dominant},\,\ref{eq:nonmember_dominant}}

\For{$\ell=1$ \textbf{to} $L$}
  \State Compute discriminative score $S_\ell$, {w.r.t.\ Eq.~\ref{eq:discriminative_score}}
\EndFor

\State $L_{\mathrm{dis}}\gets\text{top-}K\text{ layers by }S_\ell$

\ForAll{$\ell\in L_{\mathrm{dis}}$}
  \State Compute neuronal similarity $s_{\mathrm{mem}}^\ell(x)$ and $s_{\mathrm{non}}^\ell(x)$, {w.r.t.\ Eqs.~\ref{eq:ratio_mem},\,\ref{eq:ratio_non}}
\EndFor

\State Calculate $\bar s_{\mathrm{mem}},\bar s_{\mathrm{non}}$, {w.r.t.\ Eqs.~\ref{eq:agg_mem},\,\ref{eq:agg_non}}
\State Get ratio $R(x, \mathcal{M})$, {w.r.t.\ Eq.~\ref{eq:membership_score}}

\If{$R>\theta$}
  \State $A(x,M)\gets1$, {w.r.t.\ Eq.~\ref{activate_result}}
\Else
  \State $A(x,M)\gets0$
\EndIf

\Return $A(x,M)$
\end{algorithmic}
\end{algorithm}

\section{Implementation Details}
\label{app:implementation}

\subsection{Data Split Configuration}
For our experimental setup, we carefully partitioned our datasets to ensure reliable evaluation:
(i) \textit{Our Method.} We construct reference activation patterns on 200 samples, tune hyperparameter on a 200-sample validation set, and evaluate final performance on an 800-sample test set.
(ii) \textit{Probe Attack.} The probe attack uses 200 non-member samples as its training pool—half of these (100 samples) are employed to fine-tune the model. It then validates on 400 samples and reports results on the same 800-sample test set.
(iii) \textit{Other Baseline Methods.} All remaining baselines are evaluated directly on the 800-sample test set.

\subsection{Technical Implementation}

\noindent\textbf{Neuronal Activation Determination.} We register hook functions on the FFN activation functions across all Transformer layers to capture post-GELU/ReLU outputs, and declare a neuron activated whenever its output exceeds the activation threshold~$\tau$.

\noindent\textbf{Neuronal Identity Discrimination.} We construct reference activation patterns using 100 member and 100 non-member samples; experiments show this quantity sufficient for stable performance.

\noindent\textbf{Hyperparameter Selection.} After sensitivity analysis over $\tau \in [0,2]$, $\alpha \in [1.2,2.0]$, and $K \in [1,32]$, we recommend setting the activation threshold $\tau = 1.5$, the dominance coefficient $\alpha = 1.8$, and selecting $K = 5$ layers.

\noindent\textbf{Computational Resources}: Our experiment completed on a single NVIDIA A100 (40GB) GPU, with inference time approximately 30 minutes per model, varying by model size and dataset scale.

\end{document}